\documentclass{article}

\usepackage{PRIMEarxiv}

\usepackage[utf8]{inputenc} 
\usepackage[T1]{fontenc}    
\usepackage{hyperref}       
\usepackage{url}            
\usepackage{booktabs}       
\usepackage{amsfonts}       
\usepackage{nicefrac}       
\usepackage{microtype}      
\usepackage{lipsum}
\usepackage{fancyhdr}       
\usepackage{graphicx}       
\usepackage{tabularx}
\usepackage{multicol}
\usepackage{adjustbox}
\usepackage{xcolor}
\usepackage{caption}
\graphicspath{{media/}}     

\title{Mitigating Hallucinations Using Ensemble of Knowledge Graph and Vector Store in Large Language Models to Enhance Mental Health Support

\thanks{\textit{\underline{Citation}}: 
\textbf{Authors. Title. Pages.... DOI:000000/11111.}} 
}

\author{
    \textbf{Abdul Muqtadir}\\
    \textit{School of Electrical Engineering} \\
    \textit{and Computer Sciences}\\
    \textit{National University of Sciences and Technology}\\
    amuqtadr.mscs21seecs@seecs.edu.pk
  \and
    \textbf{Hafiz Syed Muhammad Bilal}\\
    \textit{School of Electrical Engineering} \\
    \textit{and Computer Sciences}\\
    \textit{National University of Sciences and Technology}\\
    bilal.ali@seecs.edu.pk
    \And
     \textbf{Ayesha Yousaf}\\
    \textit{Anatomy Department, Rawalpindi Medical} \\
    \textit{University, Rawalpindi, Pakistan}\\
    \and
    \textbf{Hafiz Farooq Ahmed}\\
    \textit{Computer Science Department, College of Computer} \\
    \textit{Sciences and Information Technology (CCSIT),} \\
    \textit{King Faisal University,}\\
    \textit{P.O. Box 400, Al-Ahsa 31982, Saudi Arabia}\\
    hfahmad@kfu.edu.sa
    \And
    \textbf{Jamil Hussain}\\
    \textit{Department of AI and Data Science,} \\
    \textit{Sejong University, Seoul,} \\
    \textit{Republic of Korea}\\
    jamil@sejong.edu.kr
}

\begin{document}
\maketitle
\begin{abstract}
This research work delves into the manifestation of hallucination within Large Language Models (LLMs) and its consequential impacts on applications within the domain of mental health. The primary objective is to discern effective strategies for curtailing hallucinatory occurrences, thereby bolstering the dependability and security of LLMs in facilitating mental health interventions such as therapy, counseling, and the dissemination of pertinent information. Through rigorous investigation and analysis, this study seeks to elucidate the underlying mechanisms precipitating hallucinations in LLMs and subsequently propose targeted interventions to alleviate their occurrence. By addressing this critical issue, the research endeavors to foster a more robust framework for the utilization of LLMs within mental health contexts, ensuring their efficacy and reliability in aiding therapeutic processes and delivering accurate information to individuals seeking mental health support.
\end{abstract}


\section{Introduction}
Mental health is an increasing concern in our fast-paced and digitally connected world \cite{prince2007no}. However, mental health services have traditionally been associated with accessibility, affordability, and stigma. Additionally, face-to-face meetings with consultants are limited in time and space. As a result, many people refuse to seek help for these problems, putting their mental health at risk. As the need for mental health support continues to increase, there is an urgent need for new developments to meet this need \cite{yonemoto2023help}. Virtual discussion techniques such as LLM’s based digital twins are being increasingly used in psychological support to help clients and patients explore, gain insight, take action, and ultimately self-heal without needing proper human support \cite{abilkaiyrkyzy2024dialogue}. However, the current virtual counseling process tends to focus on the counselor’s opinions and often ignores the patient’s behavior \cite{qiu2024psychat} \cite{liu2023chatcounselor}. This tendency leads to inappropriate or unnecessary argumentation strategies and collaborative interactions in discussion. Despite these advancements, a shortcoming of Large Language Model (LLM) content generation is hallucination, i.e. including false positive information in the response. 
In the real world, it is very difficult to make judgments about hallucinations (errors of fact or opinion in the LLM). This is because the description of semantics in the real world is still an open problem.
We define the legal world of computing as a world where failure is clearly discussed. In this world, hallucinations occur in the LLM. We draw an important conclusion from this: hallucination is an inevitable part of any dynamic LLM regardless of the model, learning algorithm, method support, or knowledge. Since the formal world is part of the real world, the results also apply to real-world LLMs \cite{deemter2024pitfalls} \cite{xu2024hallucination}. LLMs face the challenge of correctly interpreting words or concepts when the meaning is ambiguous and outside the knowledge domain of the model. This limitation poses significant challenges, particularly in critical domains such as healthcare, where accuracy is paramount \cite{ji2023survey}. Additionally, training materials contain misinformation, biases, or inaccuracies. In that case, these flaws may be reflected or amplified in the content created by these standards link which leads to outputs that may look imaginable but are often irrelevant or out of context \cite{ji2023survey}. Solving the hallucination problem in such models is difficult due to the nature of the phenomenon of these models. These models work probabilistic nature which means they may be prone to hallucinations or delusions, due to this reason they tend to deviate from the real-world scenarios and possibilities. To effectively solve this problem, there have been continuous research efforts in making knowledge bases and fine-tuning models \cite{agrawal2023can}.

\section{Related Work}
There are about 40 different types of mental health chatbots, most of which are designed to treat depression or autism. Currently,
many chatbots available combine elements of
cognitive behavioral therapy (CBT) with gaming tools to enhance the user experience, but very few chat-bots
can help manage thoughts and sentiments through a fusion of tools and techniques such as Dialectical Behavior Therapy
(DBT), evidence from CBT, and guided meditation. And they appear to be working: According to the results of a 2017 study by the Stanford University School of Medicine and the developers of the software called Woebot, the robots are quite functional in reducing depression. After just two weeks of treatment, these digital mental health support tools seem to be humane among college students. \cite{fitzpatrick2017delivering}

The utilization of LLMs in mental health services is a growing area. Researchers have proposed models that can help employees provide effective counseling or peer support. They have also created Emotional Support Conversations (ESCs) or conversations based on Hill's Helping Skills Theory \cite{hill2020helping} to create support to reduce daily stress. A major risk posed by LLMs in the domain of mental health is hallucination \cite{de2023benefits}.The traditional division of hallucination is the internal-external dichotomy\cite{huang2023survey} \cite{dziri2021neural}. Internal hallucinations occur when the LLM output is against the input. External hallucinations occur when the LLM output cannot identify the information in the input. Hallucinations are needed to be eliminated from LLM responses especially in the healthcare field in order for them to be reliable and appropriate for the one seeking their assistance and employing them instead of actual specialist. The area of large language model (LLM) hallucination mitigation has attracted a significant amount of interest lately \cite{lewis2020retrieval}. Recent comprehensive surveys \cite{ji2023survey} on the types and causes of hallucinations in natural language generation have been made available by previous works. These surveys emphasize that the probabilistic nature of LLMs, a lack of training data, or the models' attempt to patch in knowledge gaps with erroneous but plausible-sounding information can all lead to hallucinations. However, there is still room left for research on the production of integrated dialogue systems for cognitive health assistance \cite{ji2023survey}. To the best of our knowledge, this is the early research that specifically focuses on critically reviewing methods for augmenting large language models (LLMs) using combination of structured knowledge from knowledge graphs and dense vector representation from vector stores. In particular, we concentrate on addressing hallucinations in LLMs by integrating knowledge graphs and vector stores at the same time.

\section{Methodology}
Language modeling is an important task in natural language processing (NLP), which focuses on understanding language patterns and generating text. It has gained importance in recent years. Specifically, in probabilistic neural language models, the goal is to predict the probability of a sequence of words. It involves counting the probability of each coin in the sequence, including previous coins, using the chain rule to simplify the process.
The introduction of the Transformers architecture \cite{vaswani2017attention} effectively improves the probabilistic neural language model and makes it more efficient to match and identify distant areas of text. When combined with previous training such as remedial training and learning support with human language feedback (RLHF), these abstract language models have led to the formation of large-scale language models (LLMs), such as GPTLMs.

In this research, we have used open-source large language models i.e Google Gemma \cite{team2024gemma}, Mistral \cite{metz2023mistral}, and also Zypher \cite{tunstall2023zephyr} (which has been used for several healthcare-related tasks) for the generation of answers against user questions based on the mental health domain. Google Gemma has been fine-tuned by incorporating data from the above-mentioned data sources.

A retrieval-enhanced generative model such as Ensemble RAG \cite{hayden1998ensemble} \cite{lewis2020retrieval} improves the multitasking knowledge of the LLM by providing relevant information during the generative process and reduces hearing without changing the architecture of the LLM. This technique is useful for tasks that require external information by adding high-level information to the input.
  
\begin{figure*}
    \centering
    \includegraphics[width=.8\linewidth]{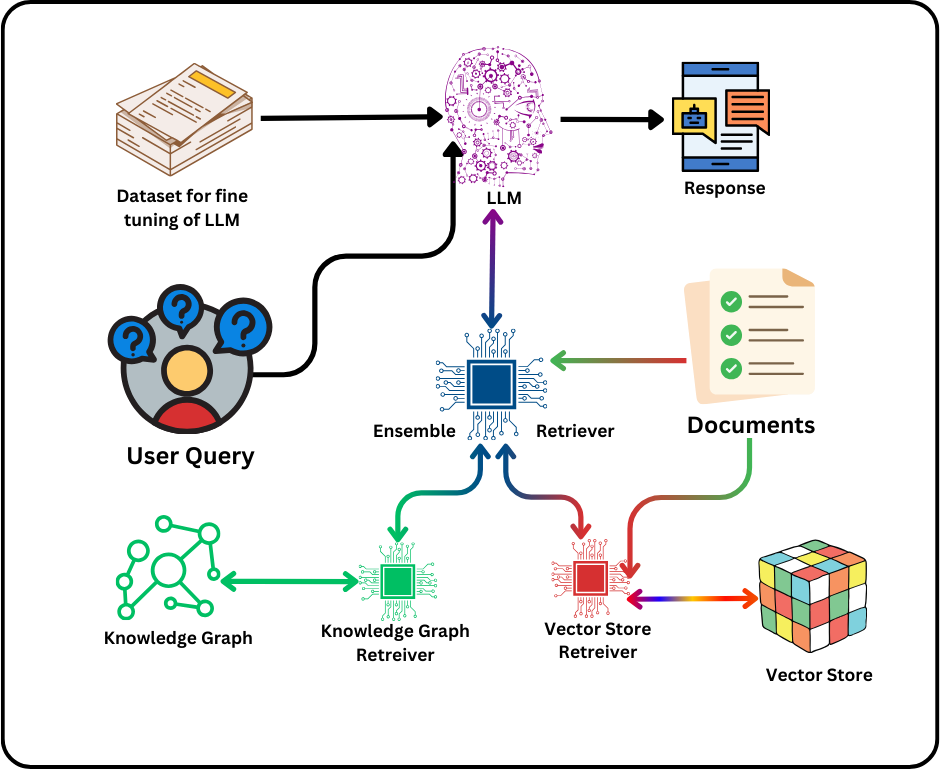}
    \caption{System architecture of the ensemble retriever framework}
    \label{fig:pipeline}
\end{figure*}

\subsection{Ensemble Retriever Framework}

The proposed solution combines the strengths of vector store retrieval and knowledge graph store retrieval to reduce hallucinations in large language models. The ensemble retriever \cite{shen2023unifier} integrates these components to provide a more accurate and contextually relevant set of information to the LLM, enhancing its output quality.

\subsection{System Architecture}
The ensemble retriever system is designed with the following components (Figure 1):
\begin{itemize}
    \item \textbf{Query Processor:} The system parses user input to understand the intent of the query and identify key terms, concepts, and entities. This may involve using techniques such as Named Entity Recognition (NER) and dependency parsing. The query is tokenized, and stop words are removed. Additional preprocessing may include lowercasing, stemming, and lemmatization to normalize the input.
\end{itemize}
\begin{itemize}
    \item \textbf{Vector Store Retriever:} The Vector Store Retrieval module is responsible for retrieving semantically relevant documents based on the vector representation of the query. All documents in the corpus are preprocessed and transformed into dense vector representations using the same embedding model as the query. The dense vector representations of the documents are indexed using ANN algorithms. This indexing enables efficient retrieval by organizing the vectors in a way that allows for quick similarity comparisons.
\end{itemize}
\begin{itemize}
    \item \textbf{Knowledge Graph Store Retriever:} The Knowledge Graph Store Retrieval module leverages structured, authoritative knowledge from the GENA (Graph for Enhanced Neuropsychiatric Analysis) knowledge graph to retrieve factually accurate information. The GENA knowledge graph \cite{dang2023gena} contains information on mental health, including symptoms, treatments, diseases, and genetic markers. Entities (nodes) in the graph represent key concepts, and relationships (edges) represent the links between them.
\end{itemize}
\begin{itemize}
    \item \textbf{Fusion Module:} The Fusion Module is the core component responsible for combining the results from both the vector store retrieval and knowledge graph store retrieval modules. It ensures that the final output is both semantically relevant and factually accurate.
    The Fusion Module aggregates the top \textit{\textbf{k}} results from the vector store retrieval and the knowledge graph store retrieval. The results from both sources are harmonized to ensure consistency and relevance. 
\end{itemize}
\begin{itemize}
    \item \textbf{Language Model:} The top-ranked results are synthesized into a coherent natural language response using a pre-trained language generation model. The language model generates grammatically correct and contextually appropriate answers based on the retrieved content.
\end{itemize}

\subsection{Vector Store Retrieval}
Vector Store Retrieval relies on embedding-based techniques to represent textual data as dense vectors in a high-dimensional space \cite{han2023comprehensive}. These vectors encode semantic meaning, allowing for effective matching and retrieval based on the similarity of their embedding rather than just keyword overlap. Some of the use full features of vector store retrieval includes semantic similarity which captures the meaning of text by embedding words, sentences, or documents into vectors that reflect their semantic content \cite{stata2000term}. Contextual Matching uses these embedding to match user queries with relevant passages, providing contextually appropriate information. Scalability which supports efficient retrieval from large datasets by leveraging vector-based indexing and search algorithms. The vector representation 
\textbf{v} of a text 
\textbf{T} can be formulated as:

\begin{equation}
v=f(T)
\end{equation}
where \textit{f} is the embedding function.
\subsubsection{Embedding Generation}
Sentence-Transformers \cite{geng2022deep}, a pre-trained model recognized for producing better sentence embeddings has been used in this research. Using the chosen model, every document in the corpus is transformed into a dense vector representation. Effective similarity searches have been seen feasible by the indexing of these vectors in the Chroma Vector Store, a high-dimensional vector store. A schema-less vector database created for artificial intelligence applications is called Chroma Vector Store. With the ability to store, retrieve, and handle vector data (embeddings) necessary for LangChain and data propagation in chat applications, it is both lightweight and powerful. Using Chroma Vector Store, vector data can be easily managed to improve the accuracy and performance of AI applications.

\subsubsection{Retrieval Process}
The user query is transformed into a dense vector using the same embedding model.The nearest neighbor search is conducted within the vector store to identify documents with the highest cosine similarity to the query vector. The top-k documents are ranked based on their similarity scores and returned for further processing. Accurate retrieval of semantically similar text can be achieved by comparing query vectors 
\textit{\textbf{q}} with indexed vectors 
\textbf{\textit{v}} 
\textit{\textsubscript{i}}
The similarity score 
\textbf{\textit{s}} between query vector 
\textbf{\textit{q}} and document vector 
\textbf{\textit{v}} 
\textit{\textsubscript{i}}
is computed as:

\begin{equation}
s(q,v_{i})=cos(q,v_{i})
\end{equation}

\subsection{Knowledge Graph Store Retrieval}
The Knowledge Graph Store Retrieval \cite{zhao2018architecture} component utilizes the GENA (Graph for Enhanced Neuropsychiatric Analysis) \cite{dang2023gena} knowledge graph to provide structured, authoritative information on psychiatric conditions. GENA supports the ensemble retriever system by offering detailed, verifiable data that enhances the factual accuracy and contextual relevance of responses, thereby mitigating hallucinations in large language models (LLMs).
Knowledge graphs (KGs) are structured representations of information where entities (nodes) are connected by relationships (edges). These graphs capture factual information in a way that is easy to query and interpret, making them ideal for providing accurate responses based on verified data.
GENA is a comprehensive knowledge graph specifically designed for mental health, integrating data on psychiatric conditions, symptoms, treatments, patient outcomes, and genetic information. The fundamental purpose is to make understanding and analysis of mental health diseases easier by fusing knowledge from gnomic studies with organized clinical data.
In the ensemble retriever system, the knowledge graph store retriever collects pertinent data from the KG, which enriches the LLM's output with factual information. This component solves the limitations of vector-based retrieval by ensuring that the generated responses are factually based on verifiable knowledge.

\subsubsection{Knowledge Graph Construction}
\textbf{Data Sources:}
Because of privacy concerns in cognitive health, most dialogue datasets for mental health support are sourced from public social platforms, crowdsourcing, and data synthesis. Dialogue datasets collected from public social platforms include the Huggingface datasets. Crowdsourcing involves high costs and time, with the Huggingface dataset serving as a typical example. Data synthesis is an effective approach in the era of large language models, often yielding a large-scale corpus. 

The dataset used for building a knowledge graph based on the mental health domain is taken from the novel graph model named “Graph of Mental-health and Nutrition Association” usually referred to as ‘GENA’ \cite{dang2023gena}. It contained 676 mental health terms with other domains related to medical health. Such a large amount of information collection lead to the generation of 2,975,076 keyword pairs between Mental Health and the other groups that were used to search for relevant healthcare abstracts.

The dataset includes documents that contain information about mental health. During the creation of knowledge this document data is also used to make use of entities from the text included in the document.

\subsubsection{Graph Building Process}
Determination and selection of entities such as conditions, treatments, symptoms, and genetic markers from source data. Determining and extracting the connections between entities by using source data and context. Create a schema that outlines the relationship and entity types that are pertinent to psychiatric analysis. Combining information from several sources, foreseeing and catering disputes with certain accuracy. Verify the graph's accuracy and completeness by doing quality checks, such as cross-referencing with reliable sources and professional evaluations.

\subsection{Query Processing}
\textbf{Query Translation:}
\begin{itemize}
    \item \textbf{Natural Language Understanding (NLU):} Convert user queries into structured queries that can be executed against GENA, involving entity recognition, intent detection, and relationship identification.
\end{itemize}
\begin{itemize}
    \item \textbf{Cypher Queries:} Depending on the graph’s underlying technology, GENA supports SPARQL (for RDF-based graphs) or Cypher (for property graphs in systems like Neo4j). We have used Neo4j for our research and have recreated the knowledge graph with all the entities.The retrieval from the knowledge graph can be represented as:
    \begin{equation}
        Results=KGQuery(G,query)
    \end{equation}
        where \textbf{\textit{G}} is the knowledge graph and query is the structured query.
\end{itemize}
\textbf{Example:} For a query such as “What genetic factors are linked to bipolar disorder?”, the system translates it into a Cypher query to find genetic markers related to "Bipolar Disorder" in GENA.

\subsection{Fusion Module}
The Fusion Module plays a critical role in the ensemble retriever system by integrating and refining the results from the vector store retriever and the knowledge graph store retriever. The primary objective is to combine the factual precision of the knowledge graph with the semantic richness of the vector store thus enhancing the benefits of both retrieval techniques which eventually helps in eliminating hallucinations. This integration reduces the likelihood of hallucinations in large language models (LLMs) by generating responses that are factually accurate and contextually relevant.

\subsubsection{Result Integration}
The fusion module integrates the results from the vector store and the knowledge graph store, aiming to produce a comprehensive and accurate response. 
The aggregated result \textbf{\textit{R}} is given by:
\begin{equation}
    R=Aggregate({R_{vector},R_{graph}})
\end{equation}
The aggregation process involves:
\begin{itemize}
    \item \textbf{Preliminary Filtering:} Both the vector store and the knowledge graph store's initial findings are evaluated using predetermined relevance and accuracy thresholds. This stage of the procedure eliminates findings that are irrelevant or of poor quality early on. This elimination ensures that the extra padding or information that can be skipped without disturbing the context is only being presented.
\end{itemize}
\begin{itemize}
    \item \textbf{Re-ranking:} By combining similarity ratings from the vector store with accuracy metrics from the knowledge graph, the results are reranked. The information that is most factually correct and contextually relevant is given priority crediting to the re-ranking procedure.
\end{itemize}
\begin{itemize}
    \item \textbf{Ensemble Strategies:}
\end{itemize}
\begin{itemize}
    \item
    \begin{enumerate}
    \textbf{Weighted Voting:}A confidence score is assigned by each retriever (knowledge graph and vector storage) to the results it has retrieved. These scores are combined by the fusion module, which weights the information from each retriever according to its degree of confidence and contextual significance. High-confidence results from both sources are provided to make a substantial contribution to the final ranking proving this strategy to be useful.
    The final score 
    \textbf{\textit{S}} can be computed as a weighted sum of individual scores:
    \begin{equation}
        S=w_{1}s_{vector}+w_{2}s_{vector}+w_{3}s_{vector}
    \end{equation}
    where \textit{\textbf{w\textsubscript{1}}}, \textit{\textbf{w\textsubscript{2}}}, and \textit{\textbf{w\textsubscript{3}}} are the weights for vector similarity, graph accuracy, and contextual relevance scores respectively.
    \end{enumerate}
\end{itemize}
\begin{itemize}
    \item 
    \begin{enumerate}
    \textbf{Decision Fusion:} The top results from each retriever are selected and cross-verified to ensure consistency and accuracy. The fusion module may choose to present results from both sources if they complement each other, or it may resolve discrepancies by preferring the more accurate or relevant information.
    \end{enumerate}
\end{itemize}
\subsubsection{Contextual Disambiguation}
\begin{itemize}
    \item \textbf{Entity Context Matching:} In order to make sure the answer conatins the intended context, the user query is cross-referenced with entities from the knowledge graph and documents that have been retrieved from the vector store. if a query specifies the term "panic attack" in relation to a mental disorder. Results pertaining to the term "panic attack" would be given precedence over those pertaining to other conditions, for example, if a query specifies the term "panic attack" in relation to a mental disorder.
\end{itemize}
\begin{itemize}
    \item \textbf{Fact-Checking:} Verification is done between the information obtained from the vector store and the structured data discovered in the knowledge graph. This stage assists in tracking down and eliminating contradictions, confirming that the finished response is factually accurate as well as semantically relevant. The knowledge graph, for instance, is used to verify the legitimacy of treatment choices for diseases that the vector store has retrieved.
\end{itemize}

\section{Case Study}
A series of medical queries were used to evaluate the system's performance, including:
\begin{itemize}
    \item “What are the symptoms and treatment options for hypertension?”
\end{itemize}
\begin{itemize}
    \item “How metformin interacts with insulin in diabetes facilitation?”
\end{itemize}
\begin{itemize}
    \item “What are the potential problems arising baecause of taking atorvastatin?”
\end{itemize}
These queries were chosen to cover a range of complexity and specificity, reflecting common patient and provider information needs.
The ensemble retriever system showed a marked improvement in accuracy and relevance compared to the baseline LLM system. Increased from 78\% to 92\%. The inclusion of knowledge graph data significantly reduced the rate of incorrect information. Enhanced contextual understanding was observed, with relevance scores improving from 70\% to 88\% as reviewed by medical experts.

For the query “How does metformin interact with insulin in diabetes management?”, Regarding diabetic drugs, the initial system offered a generic response. The ensemble retriever, on the other hand, provided an extensive rationale of the pharmacodynamics and particular interactive effects of insulin and metformin, based on clinical recommendations and organised medical data.
With the ensemble retriever, the hallucination rate decreased from 22\% in the baseline system to 5\%. The knowledge graph drastically decreased hallucinations through fact-checking and the distribution of organised, reliable material.
Occasionally, the baseline system produced answers to the question "What are the symptoms and treatment options for hypertension?" that contained inaccurate or out-of-date therapy suggestions. With the help of the knowledge graph's amended medical recommendations, the ensemble retriever regularly offered precise and up-to-date medicinal properties. 
Patients' and healthcare professionals' feedback revealed a greater level of confidence in the system's ability to respond. On a 5-point rating, satisfaction scores increased from an average of 3.8 to 4.6. Users noted that the responses were more in line with modern medical norms and practices, and they valued the increased accuracy and complexity of the material.

\begin{figure}[h!]%
    \centering
    {\includegraphics[width=14cm]{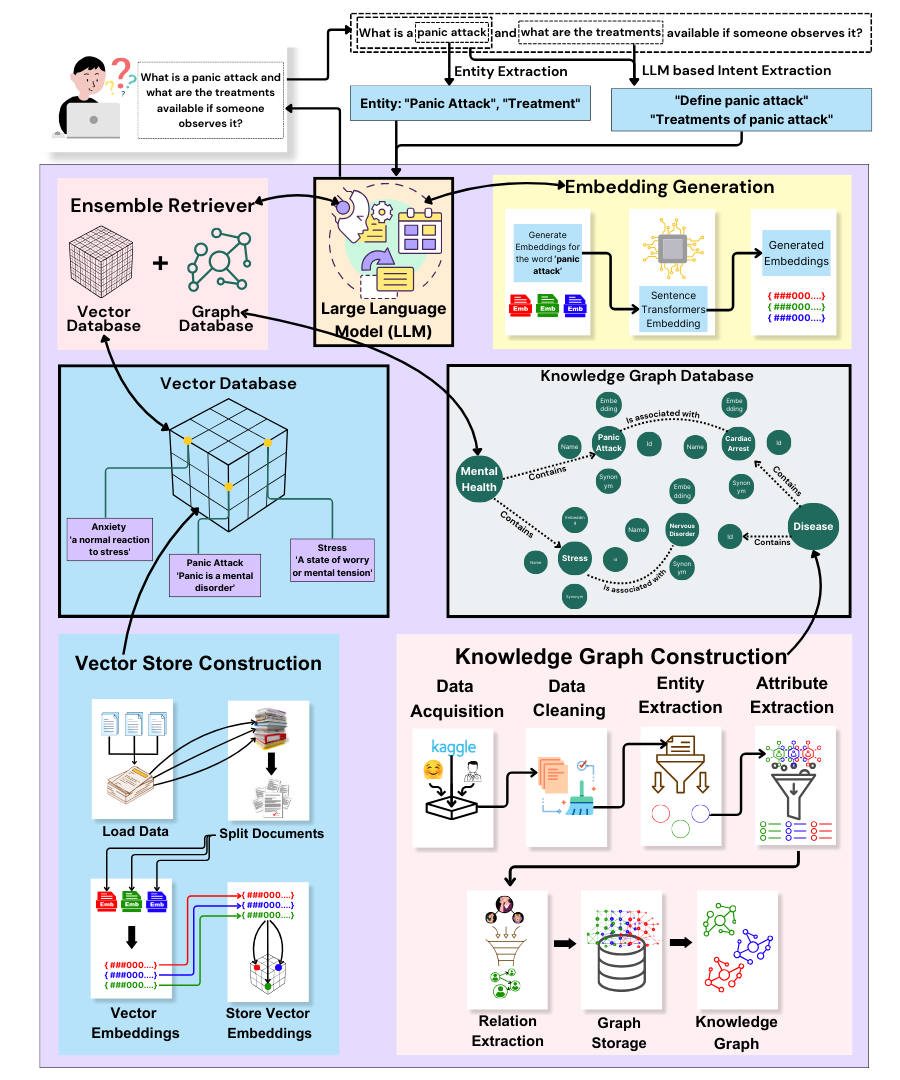} }%
    \qquad
    \label{fig:example}%
\end{figure}

\vspace{.5cm}
\newpage
\section{Performance and Evaluation:}
For the evaluation of our Ensemble RAG, we have used a popular evaluator provided by HuggingFace named evaluator. Our evaluation shows that the responses generated by ensemble RAG are better than the ones generated without ensemble evaluation either using vector-based RAG or Knowledge Graph-based RAG.

\center{\textbf{Using Google Gemma Fine-Tuned as LLM}}

\begin{minipage}{.5\textwidth}
\centering
\vspace{.20cm}
\begin{tabular}{|l|llll|}
\hline
BLEU Score    & \multicolumn{4}{l|}{0.00} \\ \hline
ROUGE-1 Score & \multicolumn{4}{l|}{0.09} \\ \hline
ROUGE-2 Score & \multicolumn{4}{l|}{0.01} \\ \hline
ROUGE-L Score & \multicolumn{4}{l|}{0.07} \\ \hline
METEOR Score  & \multicolumn{4}{l|}{0.18} \\ \hline
Overall Score & \multicolumn{4}{l|}{0.07} \\ \hline
\end{tabular}
\captionsetup{width=5cm}
\captionof{table}{Without ensemble retriever}
\end{minipage}%
\begin{minipage}{.5\textwidth}
\centering
\vspace{.20cm}
\begin{tabular}{|l|llll|}
\hline
BLEU Score    & \multicolumn{4}{l|}{0.00} \\ \hline
ROUGE-1 Score & \multicolumn{4}{l|}{0.14} \\ \hline
ROUGE-2 Score & \multicolumn{4}{l|}{0.02} \\ \hline
ROUGE-L Score & \multicolumn{4}{l|}{0.09} \\ \hline
METEOR Score  & \multicolumn{4}{l|}{0.20} \\ \hline
Overall Score & \multicolumn{4}{l|}{0.09} \\ \hline
\end{tabular}
\captionsetup{width=5cm}
\captionof{table}{With ensemble retriever}
\end{minipage}%

\begin{minipage}{.5\textwidth}
\centering
\begin{tabular}{|l|llll|}
\hline
BLEU Score    & \multicolumn{4}{l|}{0.00} \\ \hline
ROUGE-1 Score & \multicolumn{4}{l|}{0.12} \\ \hline
ROUGE-2 Score & \multicolumn{4}{l|}{0.00} \\ \hline
ROUGE-L Score & \multicolumn{4}{l|}{0.09} \\ \hline
METEOR Score  & \multicolumn{4}{l|}{0.23} \\ \hline
Overall Score & \multicolumn{4}{l|}{0.09} \\ \hline
\end{tabular}
\captionsetup{width=5cm}
\captionof{table}{With Higher Weights of Vector Store and Lower Weights of Knowledge Graph}
\end{minipage}%
\begin{minipage}{.5\textwidth}
\centering
\begin{tabular}{|l|llll|}
\hline
BLEU Score    & \multicolumn{4}{l|}{0.00} \\ \hline
ROUGE-1 Score & \multicolumn{4}{l|}{0.14} \\ \hline
ROUGE-2 Score & \multicolumn{4}{l|}{0.05} \\ \hline
ROUGE-L Score & \multicolumn{4}{l|}{0.10} \\ \hline
METEOR Score  & \multicolumn{4}{l|}{0.26} \\ \hline
Overall Score & \multicolumn{4}{l|}{0.11} \\ \hline
\end{tabular}
\captionsetup{width=5cm}
\captionof{table}{With Higher Weights of Knowledge Graph and Lower Weights of Vector Store}
\end{minipage}%

\center{\textbf{Using Mistral AI as LLM}}

\begin{minipage}{.5\textwidth}
\centering
\vspace{.20cm}
\begin{tabular}{|l|llll|}
\hline
BLEU Score    & \multicolumn{4}{l|}{0.04} \\ \hline
ROUGE-1 Score & \multicolumn{4}{l|}{0.23} \\ \hline
ROUGE-2 Score & \multicolumn{4}{l|}{0.09} \\ \hline
ROUGE-L Score & \multicolumn{4}{l|}{0.15} \\ \hline
METEOR Score  & \multicolumn{4}{l|}{0.30} \\ \hline
Overall Score & \multicolumn{4}{l|}{0.16} \\ \hline
\end{tabular}
\captionsetup{width=5cm}
\captionof{table}{Without ensemble retriever}
\end{minipage}%
\begin{minipage}{.5\textwidth}
\centering
\vspace{.20cm}
\begin{tabular}{|l|llll|}
\hline
BLEU Score    & \multicolumn{4}{l|}{0.05} \\ \hline
ROUGE-1 Score & \multicolumn{4}{l|}{0.33} \\ \hline
ROUGE-2 Score & \multicolumn{4}{l|}{0.08} \\ \hline
ROUGE-L Score & \multicolumn{4}{l|}{0.15} \\ \hline
METEOR Score  & \multicolumn{4}{l|}{0.40} \\ \hline
Overall Score & \multicolumn{4}{l|}{0.20} \\ \hline
\end{tabular}
\captionsetup{width=5cm}
\captionof{table}{With ensemble retriever}
\end{minipage}%
\begin{figure}%
    \centering
    {\includegraphics[width=7cm]{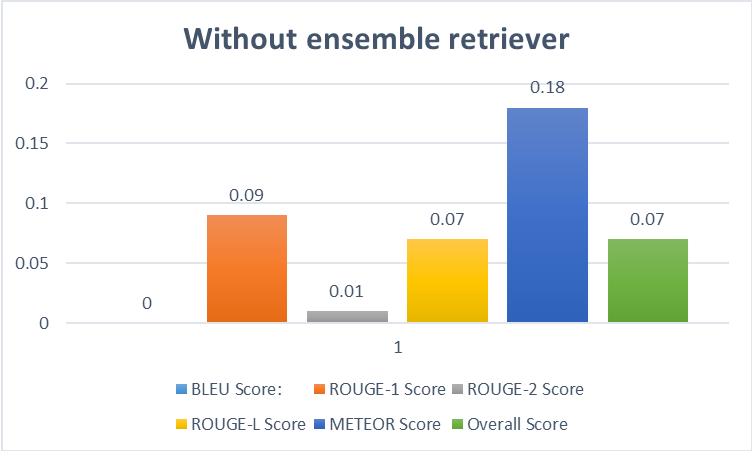} }%
    \qquad
    {\includegraphics[width=7cm]{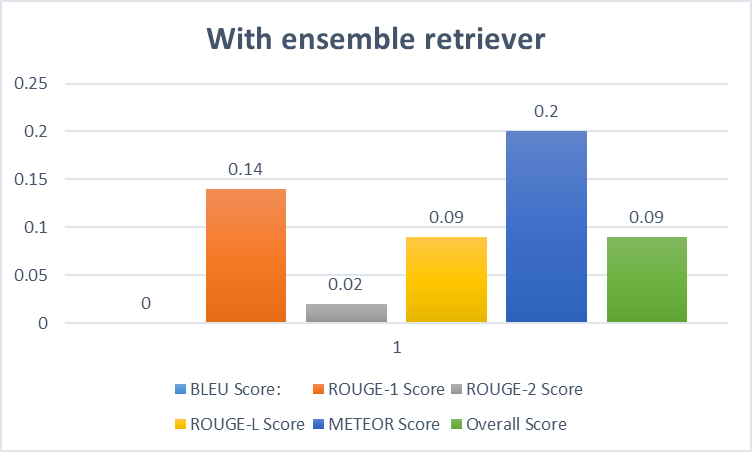} }%
    \qquad
    {\includegraphics[width=7cm]{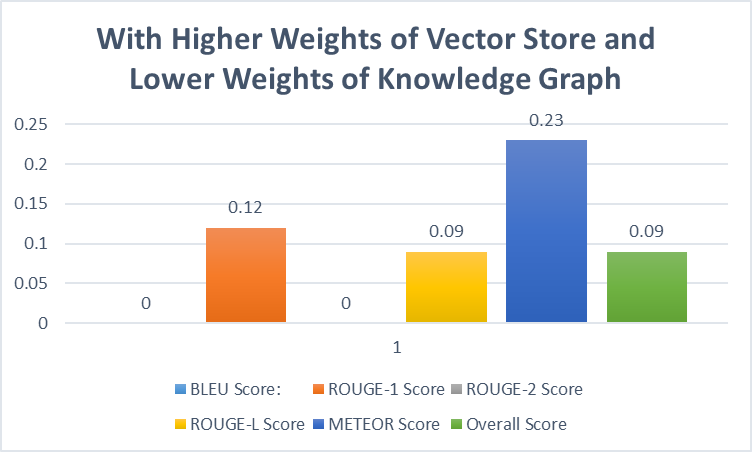} }%
    \qquad
    {\includegraphics[width=7cm]{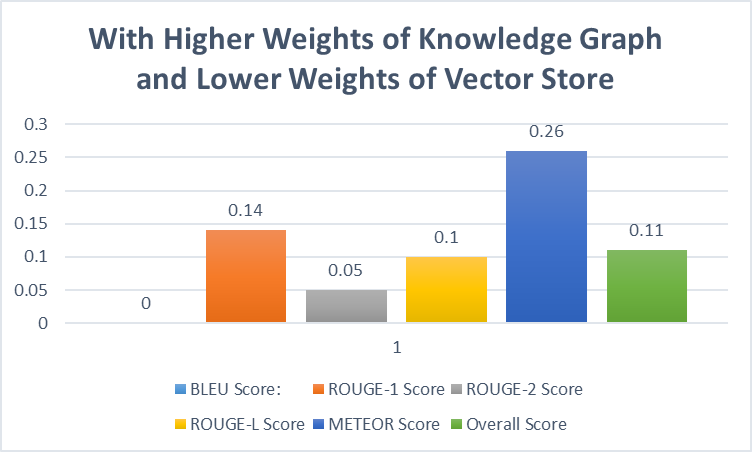} }%
    \label{fig:example}%
\end{figure}
\begin{minipage}{.5\textwidth}
\centering
\begin{tabular}{|l|llll|}
\hline
BLEU Score    & \multicolumn{4}{l|}{0.04} \\ \hline
ROUGE-1 Score & \multicolumn{4}{l|}{0.23} \\ \hline
ROUGE-2 Score & \multicolumn{4}{l|}{0.09} \\ \hline
ROUGE-L Score & \multicolumn{4}{l|}{0.15} \\ \hline
METEOR Score  & \multicolumn{4}{l|}{0.30} \\ \hline
Overall Score & \multicolumn{4}{l|}{0.16} \\ \hline
\end{tabular}
\captionsetup{width=5cm}
\captionof{table}{With Higher Weights of Vector Store and Lower Weights of Knowledge Graph}
\end{minipage}%
\begin{minipage}{.5\textwidth}
\centering
\begin{tabular}{|l|llll|}
\hline
BLEU Score    & \multicolumn{4}{l|}{0.00} \\ \hline
ROUGE-1 Score & \multicolumn{4}{l|}{0.33} \\ \hline
ROUGE-2 Score & \multicolumn{4}{l|}{0.17} \\ \hline
ROUGE-L Score & \multicolumn{4}{l|}{0.30} \\ \hline
METEOR Score  & \multicolumn{4}{l|}{0.39} \\ \hline
Overall Score & \multicolumn{4}{l|}{0.24} \\ \hline
\end{tabular}
\captionsetup{width=5cm}
\captionof{table}{With Higher Weights of Knowledge Graph and Lower Weights of Vector Store}

\end{minipage}%

\begin{figure}%
    \centering
    {\includegraphics[width=7cm]{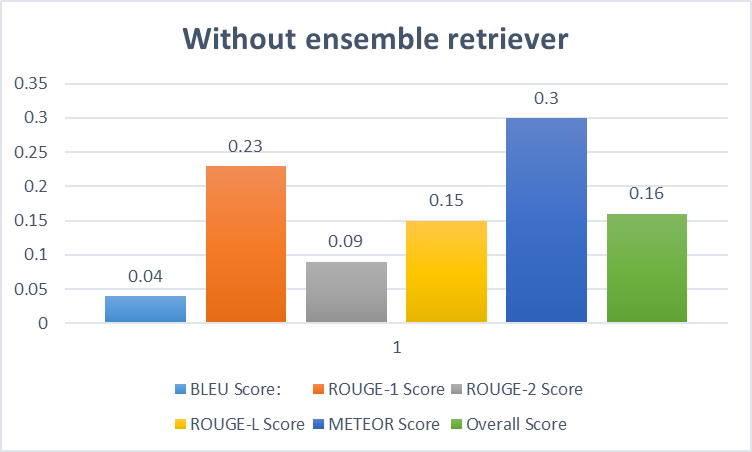} }%
    \qquad
    {\includegraphics[width=7cm]{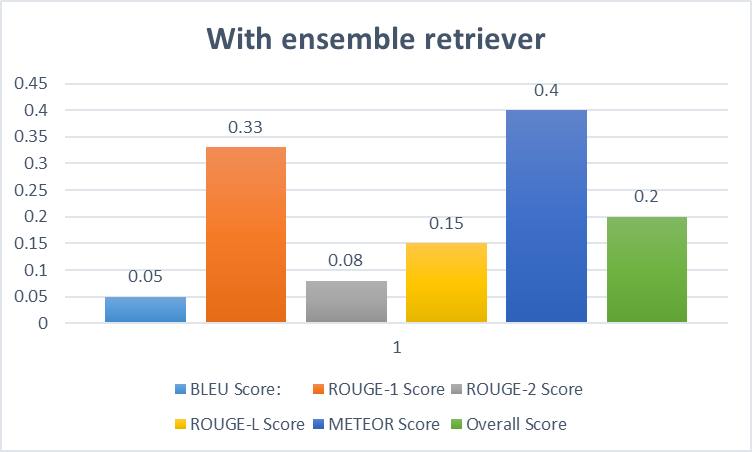} }%
    \qquad
    {\includegraphics[width=7cm]{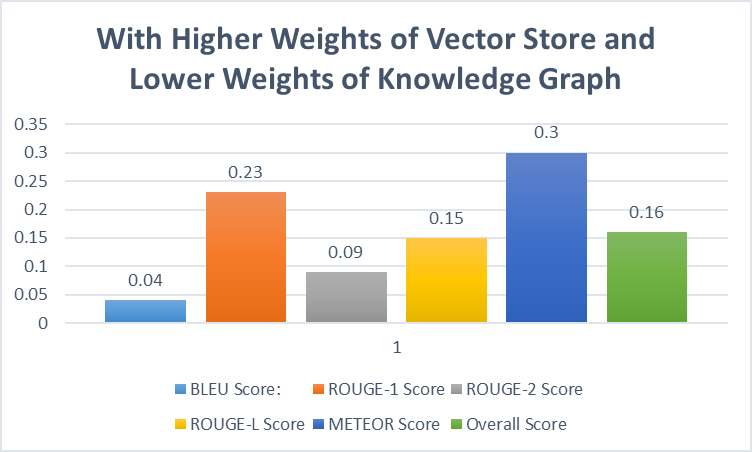} }%
    \qquad
    {\includegraphics[width=7cm]{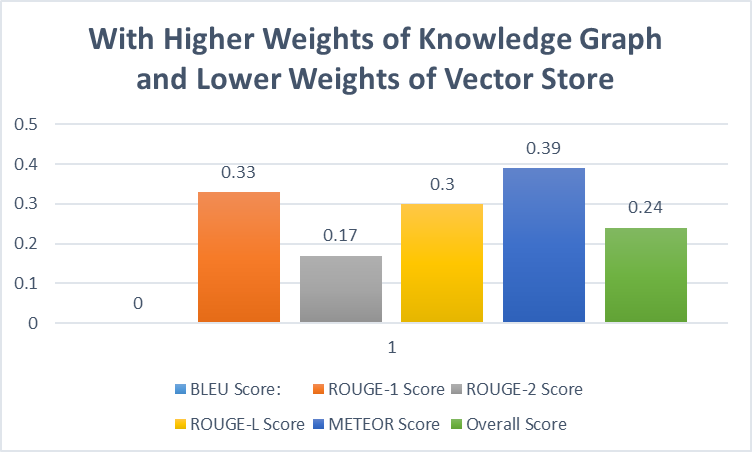} }%
    \label{fig:example}%
\end{figure}

\newpage
\center{\textbf{Using Zephyr as LLM}}

\begin{minipage}{.5\textwidth}
\centering
\vspace{.20cm}
\begin{tabular}{|l|llll|}
\hline
BLEU Score    & \multicolumn{4}{l|}{0.16} \\ \hline
ROUGE-1 Score & \multicolumn{4}{l|}{0.43} \\ \hline
ROUGE-2 Score & \multicolumn{4}{l|}{0.27} \\ \hline
ROUGE-L Score & \multicolumn{4}{l|}{0.35} \\ \hline
METEOR Score  & \multicolumn{4}{l|}{0.56} \\ \hline
Overall Score & \multicolumn{4}{l|}{0.36} \\ \hline
\end{tabular}
\captionsetup{width=5cm}
\captionof{table}{Without ensemble retriever}
\end{minipage}%
\begin{minipage}{.5\textwidth}
\centering
\vspace{.20cm}
\begin{tabular}{|l|llll|}
\hline
BLEU Score    & \multicolumn{4}{l|}{0.23} \\ \hline
ROUGE-1 Score & \multicolumn{4}{l|}{0.44} \\ \hline
ROUGE-2 Score & \multicolumn{4}{l|}{0.28} \\ \hline
ROUGE-L Score & \multicolumn{4}{l|}{0.41} \\ \hline
METEOR Score  & \multicolumn{4}{l|}{0.61} \\ \hline
Overall Score & \multicolumn{4}{l|}{0.39} \\ \hline
\end{tabular}
\captionsetup{width=5cm}
\captionof{table}{With ensemble retriever}
\end{minipage}%

\begin{minipage}{.5\textwidth}
\centering
\begin{tabular}{|l|llll|}
\hline
BLEU Score    & \multicolumn{4}{l|}{0.16} \\ \hline
ROUGE-1 Score & \multicolumn{4}{l|}{0.40} \\ \hline
ROUGE-2 Score & \multicolumn{4}{l|}{0.23} \\ \hline
ROUGE-L Score & \multicolumn{4}{l|}{0.33} \\ \hline
METEOR Score  & \multicolumn{4}{l|}{0.54} \\ \hline
Overall Score & \multicolumn{4}{l|}{0.33} \\ \hline
\end{tabular}
\captionsetup{width=5cm}
\captionof{table}{With Higher Weights of Vector Store and Lower Weights of Knowledge Graph}
\end{minipage}%
\begin{minipage}{.5\textwidth}
\centering
\begin{tabular}{|l|llll|}
\hline
BLEU Score    & \multicolumn{4}{l|}{0.23} \\ \hline
ROUGE-1 Score & \multicolumn{4}{l|}{0.44} \\ \hline
ROUGE-2 Score & \multicolumn{4}{l|}{0.28} \\ \hline
ROUGE-L Score & \multicolumn{4}{l|}{0.41} \\ \hline
METEOR Score  & \multicolumn{4}{l|}{0.61} \\ \hline
Overall Score & \multicolumn{4}{l|}{0.39} \\ \hline
\end{tabular}
\captionsetup{width=5cm}
\captionof{table}{With Higher Weights of Knowledge Graph and Lower Weights of Vector Store}
\end{minipage}%

\begin{figure}%
    \centering
    {\includegraphics[width=7cm]{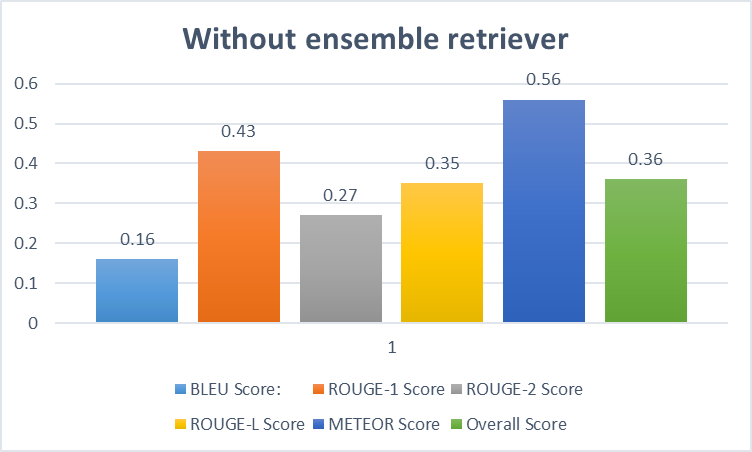} }%
    \qquad
    {\includegraphics[width=7cm]{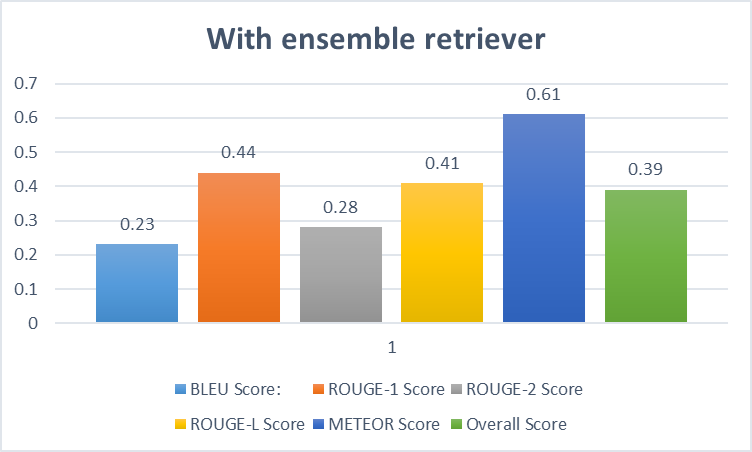} }%
    \qquad
    {\includegraphics[width=7cm]{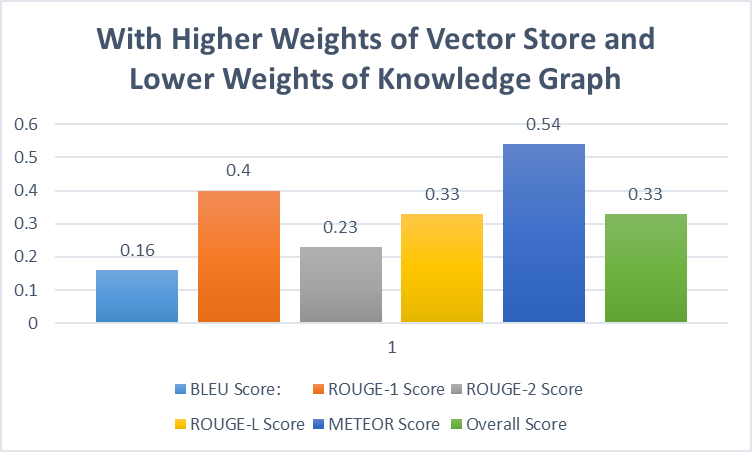} }%
    \qquad
    {\includegraphics[width=7cm]{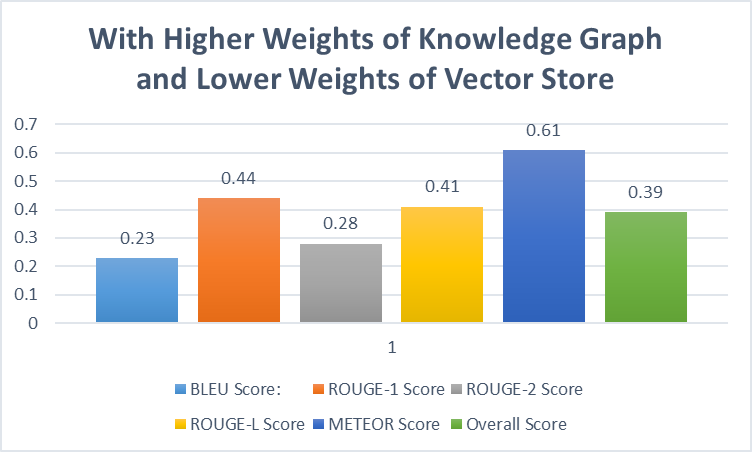} }%
    \label{fig:example}%
\end{figure}
\newpage
\begin{flushleft} From the evaluations we can clearly see that the ensemble retriever system performs better then the system relying only on single retriever. Also it can also noticed that the system is independent from the use of specific LLM's. Google Gemma was fined-tuned with the data of mental health collected from different sources but Mistral AI is still performing better then the fine-tuned LLM's. Zephyr is outperforming both of the previously used LLM's as it is the fined tuned version of Mistral AI on bio-medical datasets.\end{flushleft}

\section{Discussion }
\begin{flushleft}
The experimental results demonstrate that the proposed ensemble retriever system effectively mitigates hallucinations in large language models (LLMs). The integration of knowledge graphs significantly reduces the hallucination rate, as the structured data provides a reliable basis for fact-checking and context validation. The approach accomplishes a notable improvement in the precision and reliability of generated responses by fusing the factual precision of knowledge graph store retrieval with the semantic characteristics of vector store retrieval. The research's conclusions have a number of significant ramifications for the creation and use of LLMs in real-world settings.The decreased probability of hallucinations depicts that ensemble retrievers can increase the reliability of LLMs for jobs that demand high factual accuracy, like automated summarisation, answering questions, and creating material for delicate industries like law and healthcare.
The technology increases user happiness and trust by producing responses that are more accurate and relevant for the given context. This can be especially helpful in conversational AI systems and customer service, where users depend on the accuracy of the information given. The methodology demonstrated that it is not only efficient but also domain-adaptable to combine vector and knowledge graph storage. The system is robust as well as appropriate for a variety of knowledge-driven applications due to the fact that it can integrate many kinds of data sources.

\section{ Conclusion, Limitations and Future Work}

By utilising the mutually beneficial advantages of structured knowledge graph retrieval and sparse vector retrieval, this system architecture makes sure that the generated responses are factually correct and have a rich semantic representation. The key to reducing hallucinations in LLMs is this dual retrieval strategy. Using both vector store retrieval and knowledge graph store retrieval in the suggested ensemble retriever architecture makes LLMs significantly more accurate and reliable by minimising hallucinations, which is very useful in the mental health industry. This technique successfully blends the factual precision of knowledge graphs with the semantic richness of vector stores. This results in responses that are more relevant for the context and correct in terms of facts. The evaluations give confirmation of the advantages of this ensemble system contrasted to individual retrievers, including greater user confidence, heightened precision, and higher performance in managing difficult problems. Furthermore, the system is a viable tool for AI-powered mental health support and other applications needing a high degree of comprehension because of its ability to adapt to different datasets and sectors. This framework provides a significant improvement in the practical application of artificial intelligence (AI) by enhancing the trustworthiness of LLMs, particularly in delicate sectors like healthcare.

Despite its positive aspects, the ensemble retriever structure has some drawbacks. Its effectiveness in sectors with scarce or defective structured data may be limited by the knowledge graph's comprehensiveness and effectiveness, which have a significant impact on its performance. Large datasets in vector storage can improve semantic retrieval, but they can also introduce noise that might affect retrieval accuracy if it is not carefully regulated. Additionally, compared to independent systems which are based on a single retriever, the dual retrieval system has a greater computation cost, which leads to prolonged reaction times. This trade-off between increased accuracy and longer processing times requires adaption based on specific application needs, particularly where efficiency is critical. To improve the system's scalability and adaptability throughout a wider range of applications and domains, these limits must be resolved.

\end{flushleft}

\bibliographystyle{unsrt}  
\bibliography{references}

\begin{thebibliography}{10}

\bibitem{prince2007no}
Martin Prince, Vikram Patel, Shekhar Saxena, Mario Maj, Joanna Maselko,
  Michael~R Phillips, and Atif Rahman.
\newblock No health without mental health.
\newblock {\em The lancet}, 370(9590):859--877, 2007.

\bibitem{yonemoto2023help}
Naohiro Yonemoto and Yoshitaka Kawashima.
\newblock Help-seeking behaviors for mental health problems during the covid-19
  pandemic: A systematic review.
\newblock {\em Journal of Affective Disorders}, 323:85--100, 2023.

\bibitem{abilkaiyrkyzy2024dialogue}
Akbobek Abilkaiyrkyzy, Fedwa Laamarti, Mufeed Hamdi, and Abdulmotaleb
  El~Saddik.
\newblock Dialogue system for early mental illness detection: towards a digital
  twin solution.
\newblock {\em IEEE Access}, 2024.

\bibitem{qiu2024psychat}
Huachuan Qiu, Anqi Li, Lizhi Ma, and Zhenzhong Lan.
\newblock Psychat: A client-centric dialogue system for mental health support.
\newblock In {\em 2024 27th International Conference on Computer Supported
  Cooperative Work in Design (CSCWD)}, pages 2979--2984. IEEE, 2024.

\bibitem{liu2023chatcounselor}
June~M Liu, Donghao Li, He~Cao, Tianhe Ren, Zeyi Liao, and Jiamin Wu.
\newblock Chatcounselor: A large language models for mental health support.
\newblock {\em arXiv preprint arXiv:2309.15461}, 2023.

\bibitem{deemter2024pitfalls}
Kees~van Deemter.
\newblock The pitfalls of defining hallucination.
\newblock {\em Computational Linguistics}, 50(2):807--816, 2024.

\bibitem{xu2024hallucination}
Ziwei Xu, Sanjay Jain, and Mohan Kankanhalli.
\newblock Hallucination is inevitable: An innate limitation of large language
  models.
\newblock {\em arXiv preprint arXiv:2401.11817}, 2024.

\bibitem{ji2023survey}
Ziwei Ji, Nayeon Lee, Rita Frieske, Tiezheng Yu, Dan Su, Yan Xu, Etsuko Ishii,
  Ye~Jin Bang, Andrea Madotto, and Pascale Fung.
\newblock Survey of hallucination in natural language generation.
\newblock {\em ACM Computing Surveys}, 55(12):1--38, 2023.

\bibitem{agrawal2023can}
Garima Agrawal, Tharindu Kumarage, Zeyad Alghami, and Huan Liu.
\newblock Can knowledge graphs reduce hallucinations in llms?: A survey.
\newblock {\em arXiv preprint arXiv:2311.07914}, 2023.

\bibitem{fitzpatrick2017delivering}
Kathleen~Kara Fitzpatrick, Alison Darcy, and Molly Vierhile.
\newblock Delivering cognitive behavior therapy to young adults with symptoms
  of depression and anxiety using a fully automated conversational agent
  (woebot): a randomized controlled trial.
\newblock {\em JMIR mental health}, 4(2):e7785, 2017.

\bibitem{hill2020helping}
Clara~E Hill.
\newblock {\em Helping skills: Facilitating exploration, insight, and action}.
\newblock American Psychological Association, 2020.

\bibitem{de2023benefits}
Munmun De~Choudhury, Sachin~R Pendse, and Neha Kumar.
\newblock Benefits and harms of large language models in digital mental health.
\newblock {\em arXiv preprint arXiv:2311.14693}, 2023.

\bibitem{huang2023survey}
Lei Huang, Weijiang Yu, Weitao Ma, Weihong Zhong, Zhangyin Feng, Haotian Wang,
  Qianglong Chen, Weihua Peng, Xiaocheng Feng, Bing Qin, et~al.
\newblock A survey on hallucination in large language models: Principles,
  taxonomy, challenges, and open questions.(2023).
\newblock {\em arXiv preprint arXiv:2311.05232}, 2023.

\bibitem{dziri2021neural}
Nouha Dziri, Andrea Madotto, Osmar Za{\"\i}ane, and Avishek~Joey Bose.
\newblock Neural path hunter: Reducing hallucination in dialogue systems via
  path grounding.
\newblock {\em arXiv preprint arXiv:2104.08455}, 2021.

\bibitem{lewis2020retrieval}
Patrick Lewis, Ethan Perez, Aleksandra Piktus, Fabio Petroni, Vladimir
  Karpukhin, Naman Goyal, Heinrich K{\"u}ttler, Mike Lewis, Wen-tau Yih, Tim
  Rockt{\"a}schel, et~al.
\newblock Retrieval-augmented generation for knowledge-intensive nlp tasks.
\newblock {\em Advances in Neural Information Processing Systems},
  33:9459--9474, 2020.

\bibitem{vaswani2017attention}
Ashish Vaswani, Noam Shazeer, Niki Parmar, Jakob Uszkoreit, Llion Jones,
  Aidan~N Gomez, {\L}ukasz Kaiser, and Illia Polosukhin.
\newblock Attention is all you need.
\newblock {\em Advances in neural information processing systems}, 30, 2017.

\bibitem{team2024gemma}
Gemma Team, Thomas Mesnard, Cassidy Hardin, Robert Dadashi, Surya Bhupatiraju,
  Shreya Pathak, Laurent Sifre, Morgane Rivi{\`e}re, Mihir~Sanjay Kale,
  Juliette Love, et~al.
\newblock Gemma: Open models based on gemini research and technology.
\newblock {\em arXiv preprint arXiv:2403.08295}, 2024.

\bibitem{metz2023mistral}
Cade Metz.
\newblock Mistral, ai start-up with open-source ethos, is valued at 2 billion.
\newblock {\em The New York Times}, pages B2--B2, 2023.

\bibitem{tunstall2023zephyr}
Lewis Tunstall, Edward Beeching, Nathan Lambert, Nazneen Rajani, Kashif Rasul,
  Younes Belkada, Shengyi Huang, Leandro von Werra, Cl{\'e}mentine Fourrier,
  Nathan Habib, et~al.
\newblock Zephyr: Direct distillation of lm alignment.
\newblock {\em arXiv preprint arXiv:2310.16944}, 2023.

\bibitem{hayden1998ensemble}
Mark~Garland Hayden.
\newblock {\em The ensemble system}.
\newblock Cornell University, 1998.

\bibitem{shen2023unifier}
Tao Shen, Xiubo Geng, Chongyang Tao, Can Xu, Guodong Long, Kai Zhang, and Daxin
  Jiang.
\newblock Unifier: A unified retriever for large-scale retrieval.
\newblock In {\em Proceedings of the 29th ACM SIGKDD Conference on Knowledge
  Discovery and Data Mining}, pages 4787--4799, 2023.

\bibitem{dang2023gena}
Linh~D Dang, Uyen~TP Phan, and Nhung~TH Nguyen.
\newblock Gena: a knowledge graph for nutrition and mental health.
\newblock {\em Journal of Biomedical Informatics}, 145:104460, 2023.

\bibitem{han2023comprehensive}
Yikun Han, Chunjiang Liu, and Pengfei Wang.
\newblock A comprehensive survey on vector database: Storage and retrieval
  technique, challenge.
\newblock {\em arXiv preprint arXiv:2310.11703}, 2023.

\bibitem{stata2000term}
Raymie Stata, Krishna Bharat, and Farzin Maghoul.
\newblock The term vector database: fast access to indexing terms for web
  pages.
\newblock {\em Computer Networks}, 33(1-6):247--255, 2000.

\bibitem{geng2022deep}
Qitao Geng, Runtao Yang, and Lina Zhang.
\newblock A deep learning framework for enhancer prediction using word
  embedding and sequence generation.
\newblock {\em Biophysical Chemistry}, 286:106822, 2022.

\bibitem{zhao2018architecture}
Zhanfang Zhao, Sung-Kook Han, and In-Mi So.
\newblock Architecture of knowledge graph construction techniques.
\newblock {\em International Journal of Pure and Applied Mathematics},
  118(19):1869--1883, 2018.

\end{thebibliography}

\end{document}